\documentclass[manuscript]{geophysics}
\usepackage{graphicx}
\usepackage{wrapfig}
\usepackage{graphicx}
\usepackage{float}
\usepackage{indentfirst}
\usepackage[T1]{fontenc} 
\usepackage{fourier} 
\usepackage[english]{babel} 
\usepackage{amsmath,amsfonts,amsthm, amssymb} 
\usepackage{lipsum} 

\usepackage{caption}
\usepackage{graphicx}
\usepackage{CJK}
\usepackage{indentfirst}

\usepackage{float}
\usepackage[ruled]{algorithm2e}
\usepackage{wrapfig,lipsum,booktabs} 
\usepackage{multirow}
\usepackage[ruled]{algorithm2e}
\usepackage{algpseudocode}

\usepackage{commath}

\usepackage{mathtools}
\usepackage{setspace}
\usepackage[T1]{fontenc}

\usepackage[dvipsnames]{xcolor}

\usepackage{blindtext} 

\usepackage[]{hyperref}  

\usepackage{algorithm2e}
\SetKwInOut{Parameter}{Parameters }
\SetKwInOut{PredefinedFunction}{Methods}
\SetKwInOut{Input}{Input}
\SetKwInOut{Output}{Output}

\newlength\mylen
\newcommand\myinput[1]{%
  \settowidth\mylen{\Input{}}%
  \setlength\hangindent{\mylen}%
  \hspace*{\mylen}#1\\}

\usepackage{fancyhdr} 
\begin{document}

\title{Attenuating Random Noise in Seismic Data by a Deep Learning Approach}

\renewcommand{\thefootnote}{\fnsymbol{footnote}} 

\address{
\footnotemark[1]Texas A\&M University, 
3112 TAMU, 
College Station, TX-77843 \\
\footnotemark[2]Anadarko Petroleum Corp.,
1201 Lake Robbin's Dr,
The Woodlands, TX-77380\\
\footnotemark[3]University of Houston, 
4800 Calhoun Rd, 
Houston, TX-77004,\\ 
\footnotemark[4] Authors contributed equally to this manuscript}
\author{Xing Zhao\footnotemark[1],
Ping Lu\footnotemark[2]\footnotemark[4],
Yanyan Zhang\footnotemark[2]\footnotemark[4],
Jianxiong Chen\footnotemark[2], and
Xiaoyang Li\footnotemark[3]}

\footer{Example}
\lefthead{Zhao \textit{et al.}}
\righthead{Seismic Noise Attenuation by DL}

\maketitle

\begin{abstract}
  In the geophysical field, seismic noise attenuation has been considered as a critical and long-standing problem, especially for the pre-stack data processing. Here, we propose a model to leverage the deep-learning model for this task. Rather than directly applying an existing de-noising model from ordinary images to the seismic data, we have designed a particular deep-learning model, based on residual neural networks. It is named as N2N-Seismic, which has a strong ability to recover the seismic signals back to intact condition with the preservation of primary signals. The proposed model, achieving with a great success in attenuating noise, has been tested on two different seismic datasets. Several metrics show that our method outperforms conventional approaches in terms of Signal-to-Noise-Ratio, Mean-Squared-Error, Phase Spectrum, etc. Moreover, robust tests in terms of effectively removing random noise from any dataset with strong and weak noises have been extensively scrutinized in making sure that the proposed model is able to maintain a good level of adaptation while dealing with large variations of noise characteristics and intensities.
\end{abstract}

\section{Introduction}
A successful separation between true reflection signals and unwanted noise is a long-standing problem in the area of seismic data processing, and greatly affects the fidelity of subsequent seismic imaging (e.g., \cite{claerbout1985fundamentals} and \cite{dai2012multi}), geophysical inversion, like amplitude-variation-with-offset (AVO) inversion (e.g., \cite{buland2003bayesian} and \cite{li2014multicomponent}), full waveform inversion (e.g., \cite{pratt1999seismic}, \cite{chen2016detecting, chen2018improved}), and geological interpretation (e.g., \cite{brown2011interpretation}). Seismic data is inevitably affected by different types of noise. The existence of noise in the pre-stack seismic data affects the amplitude information, and thus causes unreliable inversion results. For post-stack seismic data, the existence of noise hurts the ability of interpretation, which directly affects the modelling of subsurface reservoirs.


In fact, seismic noise attenuation has gone through a long history of development. This history could be traced back to the simplest method - stacking the seismic data along the offset direction (\cite{mayne1962common}). Soubaras introduced the F-X projection filtering method in (\cite{soubaras1995prestack}). Spitz proposed a prediction error filtering method for recognizing coherent signal in the F-X domain (\cite{spitz1999pattern}). F-X deconvolution (\cite{canales1984random, naghizadeh2012multicomponent}) introduced by Canales becomes the most wide-used method for random noise attenuation recently. In addition, Zhou \textit{et al.} classify seismic noise attenuation methods into different categories based on their theories: sparse transform approach transforms seismic data to a sparse domain by applying a soft thresholding to the coefficients, and then transforms  the sparse coefficients into time-space domain, e.g., \cite{donoho1994ideal, pratt1998gauss, naghizadeh2012seismic, candes2006fast, herrmann2007non}; decomposition-based approaches decompose the corrupt seismic data into components and chooses the principal components for signal representation, e.g., \cite{chen2014random, gan2016improved, bekara2007local, fomel2013seismic, wu2018data}; and rank-reduction based approaches use low-rank seismic data during the data rearrangement processing, e.g., \cite{vautard1992singular, trickett2008f, oropeza2011simultaneous, cheng2015separation, anvari2017seismic}. In summary, most of these conventional methods utilize signal features, e.g., wave-number and frequency, and domain transformation, to attenuate seismic noise.


On the other hand, noise attenuation for images using machine learning technology has achieved great success in the computer vision field. Deep learning (DL) on denoising task for images has been developed in the past decades (\cite{jain2009natural, rabie2005robust, xie2012image}), and many research works indicate that \cite{burger2012image} achieve a giant leap in this field. Convolutional Neural Network (CNN) has an objective to train the network through learning lower dimension representations of the image features. Taking a benefit from the CNN, deep residual network (ResNet: \cite{he2016deep}), and batch normalization (\cite{ioffe2015batch}), Zhang \textit{et al.} proposed the de-noising CNN model (DnCNN), and it outperforms the traditional non-CNN based methods(\cite{zhang2017beyond}). After a short while, CNN based techniques on noise attenuation have been widely and continually developed into many variants. Recently, Noise2Noise (\cite{lehtinen2018noise2noise}) model was introduced for the noise attenuation task without providing ground-truth information. CBDNet (\cite{guo2019toward}) is comprised of two sub-networks, i.e., noise estimation and non-blind denoising, and it achieves state-of-the-art results in terms of both quantitative metrics and visual quality. Similarly, FFDNet (\cite{zhang2018ffdnet}), RED30 (\cite{mao2016image}), BM3D-Net (\cite{yang2017bm3d}), and CS-DIP (\cite{van2018compressed}) also achieved impressive performances.
    
However, directly applying these methods onto seismic data may not be effective, since geophysical domain requires not only the visual quality of the seismic image but also the recovery quality of seismic signals. For example, for training purpose, the conventional neural network would decrease the loss value, e.g., L1 (\textit{Laplace}) loss or L2 (\textit{Gauss}) loss, and make the predicted value converge to a certain level. As a result, de-noised images may have lower sharpness and looks smoother among adjacent pixels. Such techniques may not fit the seismic noise attenuation since the priority of the noise attenuation is to keep the phase and amplitude spectrum of the valid signal as intact as possible.

In this manuscript, we bring the state-of-the-art techniques of noise attenuation from computer version field into geoscience field and make a variant of the technique to fit the requirements in the geophysical domain for seismic noise attenuation. We will firstly introduce our model frame, and secondly apply our model onto two cases: the synthetic wedge dataset, and SEAM Phase I data (\cite{fehler2011seam}), following by results analysis respectively. In summary, deep neural network models proposed in this manuscript could extract the principal components of seismic data to attenuate strong noises and outliers, eventually to recover amplitudes and keep the original phase of the primary signals, without harms on the primary signals.


\section{Methodology}
\label{sec:methods}

In this section, we would like to borrow the ideas from pioneering deep learning techniques for image processing and introduce our de-noising model, N2N-Seismic, for seismic noise attenuation task. Specifically, we will first introduce some fundamental concepts of deep learning techniques for image denoising and a state-of-the-art DL-based model, Noise2Noise; secondly, we will discuss why they could not be effectively applied on seismic data processing. Then, we will introduce our DL-based solution, N2N-Seismic, specifically designed for seismic image de-noising.

\subsection{The Noise2Noise Model}
\label{sec:method_1}

Traditional methods monitor the performance of deep learning model for noise attenuation tasks from the difference between generated (or de-noised) image and the clear ground truth.  However, in recent research, \cite{lehtinen2018noise2noise} introduced the Noise2Noise model, which could learn to turn corrupted images into good ones by only looking at noised images. Noise2Noise training attempts to learn a mapping between pairs of independently degraded versions of the same training image, i.e., $(s + n, s + n')$, that incorporates the same signal $s$, but with independently random noise $n$ and $n'$. Naturally, conventional neural networks do not have the ability to learn how to map one noisy image from another noisy image. However, networks trained on this training task enable to produce results that close to the same performance as traditionally trained networks that do have access to the ground truth images. In cases where ground-truth data is physically unobtainable, N2N still could perform the tasks through training. The similar idea has been applied to its variants, e.g., Noise2Void (\cite{krull2019noise2void}).

Such deep-learning-based methods effectively work on the traditional image noise attenuation. However, in the geophysical field, experts care more about physical metrics, e.g., amplitude, phase Spectrum, amplitude spectrum, etc. Using only the Noise2Noise model on seismic data may obtain results without making physical sense. It can be found by an observation that results from the traditional Noise2Noise model, $F_g(S|\theta)$, may always flatten signals in the de-noised results, which implies the phase information of seismic is lost. It is due to the loss function, L2 Loss, which only calculates the (power of) absolute distance between the predicted value and the true value, rather than the fluctuation of the signal. As a result, if the noise level is high, de-noised signal would become flatten when the amplitude of the input signal is extremely low, for the purpose of minimizing the loss value.

For an ordinary image, it is acceptable since smooth transition would bring better visual inspection. However, for seismic data, these smooth transitions would lose important geological information, such as causing the phase distortion of seismic data, which is harmful the follow-up processing and analysis. Furthermore, unlike ordinary images where pixels evenly distribute in (0, 255) for RGB images, seismic data have severe outliers with extremely high/low amplitudes. Such outlier signals would affect the de-noising performance from the traditional Noise2Noise model. Therefore, in this paper, what is emphasized on is to keep the signal phase and amplitude as accurate as possible in both the time and frequency domains. We have developed a variant of the traditional Noise2Noise model, which is named as N2N-Seismic, specifically targeting on solving seismic noise attenuation tasks.

\subsection{N2N-Seismic: A Variant N2N model for Seismic Noise}
\label{sec:method_2}

\subsubsection{Model Design Part (1): Applying ResNet}

A Convolutional Neural Network (CNN) has been widely used in image processing, e.g. image classification  (\cite{krizhevsky2012imagenet}), face recognition  (\cite{lawrence1997face}). Residual Network (ResNet: \cite{he2016deep}), a deeper version of CNN, solved the problem that deeper network would cause a higher training/testing loss. ResNet splits the original mapping $x \Rightarrow H(x)$ to two parts: 
\begin{equation}
    x \Rightarrow f(x) \textrm{(residual mapping)}
\end{equation}
and 
\begin{equation}
    x  + f(x) \Rightarrow H(x).
\end{equation} 

where $x$ denotes the original identity, $f(x)$ denotes the residual mapping, and $H(x)$ denotes the final mapping. In this way, the problem of vanishing or exploding gradient and degradation of traditional stacked deep CNN would be eliminated. In the recent years, as one of the applications, ResNet has been widely used into image de-noising, such as the super-resolution residual network (SRResNet: \cite{ledig2017photo}; EDSR+: \cite{lim2017enhanced}). In this paper, we will design our deep learning model based on the ResNet techniques. 

\subsubsection{Model Design Part (2): Model Structure}

\begin{figure*}
    \centering
    \includegraphics[width =\textwidth]{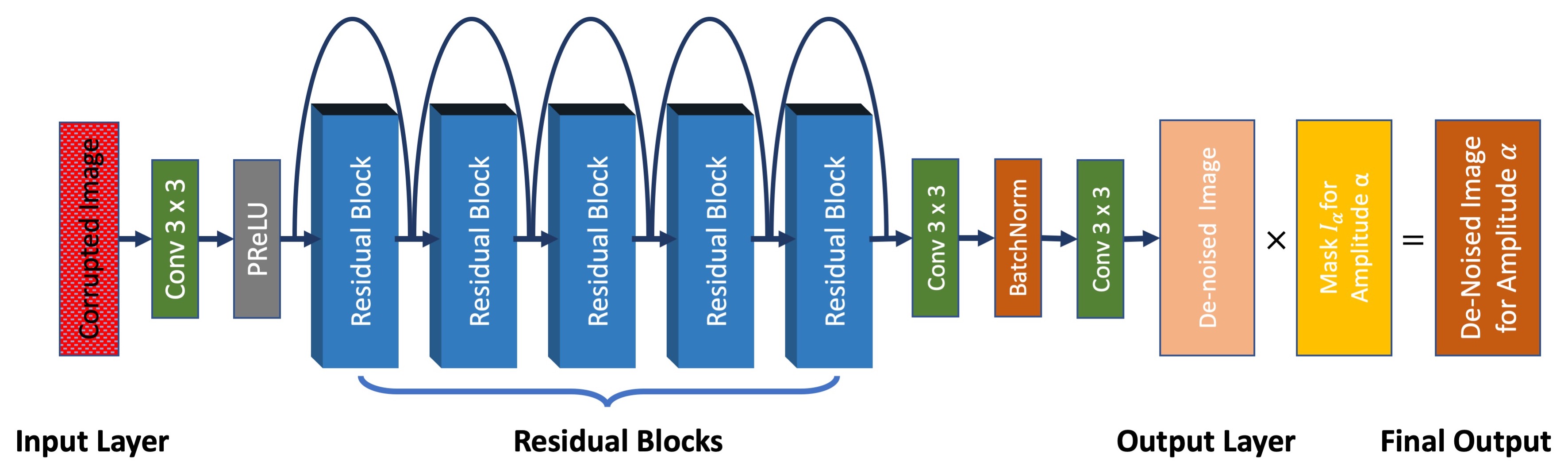}
    \vspace{-30pt}
    \caption{N2N-Seismic Network Architecture. The model takes a corrupted image as input and returns de-noised image.}
    \label{fig:model_structure}
\end{figure*}

\begin{wrapfigure}{r}{4cm}
\includegraphics[width=3.5cm]{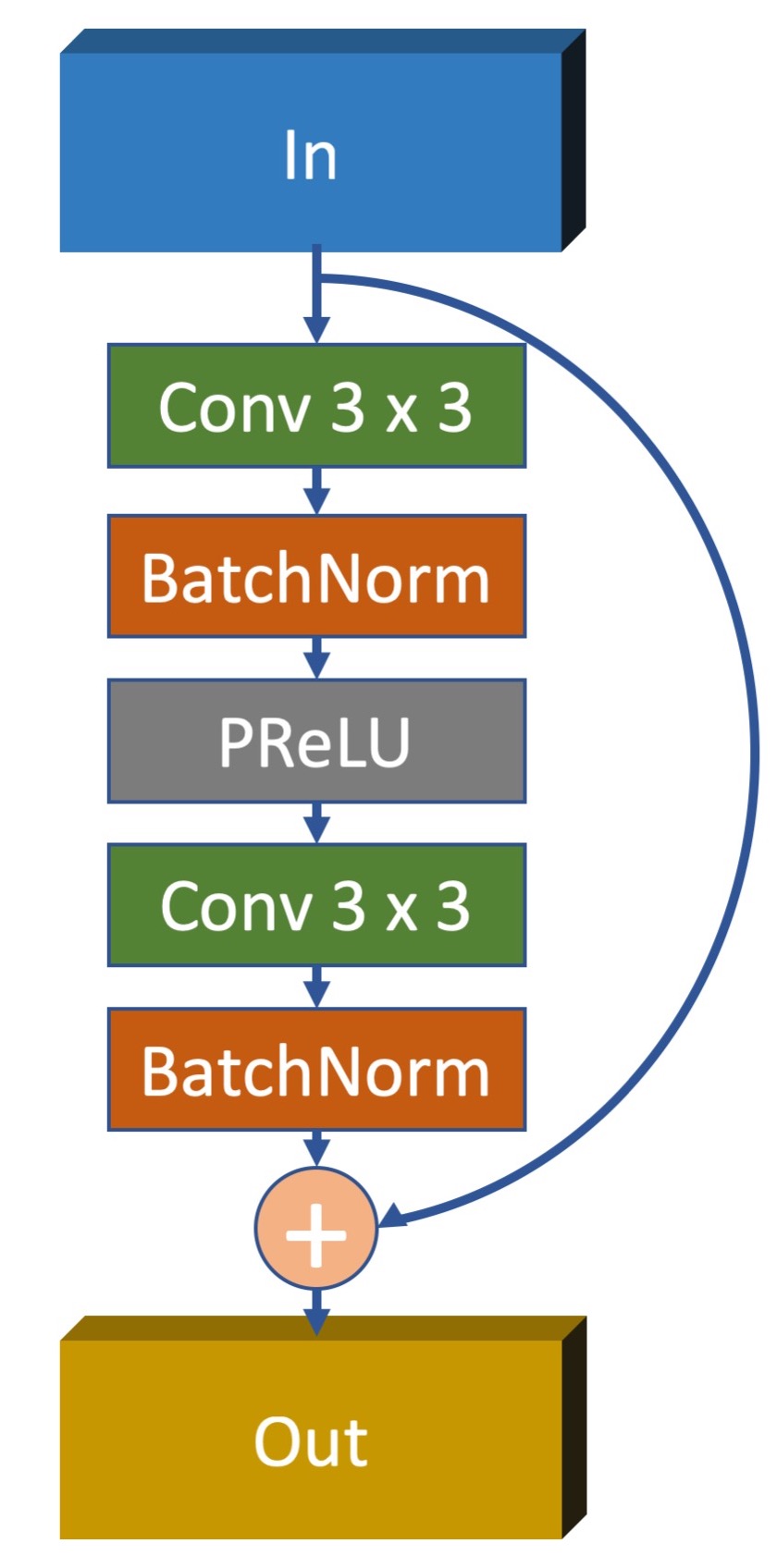}
\vspace{-30pt}
\caption{A residual unit with 5 internal layers.}
\label{fig:residual_block}
\end{wrapfigure}

Figure \ref{fig:model_structure} shows the structure of N2N-Seismic model. N2N-Seismic takes the corrupted images as inputs and return the de-noised images. Inside of the model, we connect a bunch of residual units. We use the shortcut connection to link the input and output of each residual unit together (see Figure \ref{fig:residual_block}). In each residual unit, we use the same $3\times3$ convolutional layer followed by a batch normalization layer to expedite the convergence and avoid overfitting. For the activation function, as SRResNet, we employ Parametric ReLU, instead of ReLU used in the traditional ResNet. 

Our random noise attenuation model would generate de-noised image  $F_{g}(S|\theta)$ using the given corrupted signal $S$ as input. Then, we borrow the idea from Noise2Noise model, and compare our de-noised image with another corrupted image $\hat{S}$. Here, we use the L2 loss function for training purpose, which minimizes the mean of all the squared differences between $\hat{S}$ and the predicted $F_{g}(S|\theta)$. Given a predicted image matrix with dimension $M \times N$, $F(S|\theta)^{M \times N}$, and the target ground truth image matrix, $\hat{S}^{M \times N}$, we calculate the loss as follows:

\begin{equation}
    loss = \sum_{i=1}^{M} \sum_{j=1}^{N}  \left| \left|F(S|\theta)_{i,j} - \hat{S}_{i,j}\right|\right|^2.
    \label{equation:noise2noise_loss}
\end{equation}

Back-propagation training processes would learn the optimal parameters, $\theta$, of the neural network until $loss$ converges. Based on the loss function, we now use the pair of predicted value $F_g(x|\theta)$ and the ground truth value $\hat{S}$ to tune the parameters $\theta$ to minimize the pixel-wise loss:

\begin{equation}
    arg \min \limits_{\theta}\sum_{i=1}^{M} \sum_{j=1}^{N}  \left| \left|F_g(S|\theta)_{i,j} - \hat{S}_{i,j}\right|\right|^2.
\end{equation}

\subsubsection{Model Design Part (3): A Follow-up Step for Seismic Data}

Since unlike ordinary image with normally distributed value, the seismic data may contain some portions with extremely low or high amplitude. As we discussed, traditional deep-learning based models would focus more on these outliers since they contribute more on the loss function. At the meantime, the other portions with low absolute amplitude would be predicted close to the mean of data. As a result, de-noised signals with low absolute amplitude would become flatten, we call such phenomenon as  \textit{signal information loss}.

To solve the information loss problem on the signals with low absolute amplitude, we introduce our iterative follow-up steps, Clip \& De-noise, after we have the trained model.

First, we clip the corrupted seismic input based on their absolute amplitude. Specifically, for noised seismic data $x$ with different level of absolute amplitude $\alpha_1, \alpha_2, ..., \alpha_t$, we clip $x$ by function $C(x, \alpha)$, we clip $x$ by function $C(x,\alpha_k)$ using the following criteria:

\begin{equation}
C(x_{i,j}, \alpha_k) =\left\{
\begin{array}{rcl}
- \alpha_k       &      & {x_{i,j}      <      -\alpha_k}\\
\alpha_k    &      & {x_{i,j} > \alpha_k}\\
x_{i,j}    &      & otherwise
\end{array} \right. 
\end{equation}

We also mark clipped value using a binary matrix, $I$, where $I_{i,j} = 1$ if $x_{i,j}$ has been clipped by $C(x, \alpha_k)$. Then, we apply our trained model on a different level of amplitude on the decreasing order, which is calculated by:

\begin{equation}
    \begin{aligned}
         F_{S}\left(x\left|\right.\theta\right) = &F_I\left(C\left(x, \alpha_t\right)\left|\right.\theta\right) \odot I_t\\
        & + F_I\left(C\left(x, \alpha_{t-1}\right)\left|\right.\theta\right) \odot I_{t-1}\\
        & + \cdots\\
        & +F_I\left(C\left(x, \alpha_2\right)\left|\right.\theta\right) \odot I_2\\
        & + F_I\left(C\left(x, \alpha_1\right)\left|\right.\theta\right) \odot \left(\mathbb{I} - I_t- I_{t-1} - \cdots - I_2\right)
    \end{aligned}  
    \label{equation:clipping_and_applying}
\end{equation}

where $\mathbb{I}$ denotes an all-ones matrix, and $\odot$ denotes element-wise product. $F_{S}(x|\theta)$ needs to call $t-1$ times $F_I(C(x, \alpha_{t-1})$ internally following by the final step to process the residual pixels out of $I_2, I_3, \cdots, I_t$. The output after this iterative processing would be the final de-noised seismic data. The full algorithm is listed in Algorithm \ref{alg:N2N}.

\begin{algorithm}[H]
\label{alg:N2N}
\SetAlgoLined
\Input{Corrupted image $x^{m \times n}$;  Number of amplitude range $t$;  }
\myinput{Absolute amplitude range $[\alpha_1, \alpha_2, \dots \alpha_t ]$; Binary musk $[I_1, I_2, \dots \I_t ]$}

\Parameter{Pre-trained $\theta$ for N2N-Image $F_I$; }

\PredefinedFunction{Clip function $C(x, \alpha)$, De-noise function $F_I(x|\theta)$}

\Output{De-noised image $x'$}
$x' = [0]^{m\times n}$\tcp*{Initialize Output data}

 $x_{all} = F_I\left(x\left|\right.\theta\right)$\tcp*{De-noise for all range data}
 $x' += x_{all} \odot I_t$\tcp*{keep results for those will be clipped in $\alpha_{t}$.}
 
\For{$i\gets t$ \KwTo $2$}{
    $x_c = C\left(x, \alpha_i\right)$\tcp*{Clip $x$ in the range $[-\alpha_i, \alpha_i]$}
    $x_c' = F_I\left(x_c\left|\right.\theta\right)$\tcp*{Noise attenuation for current amplitude range $\alpha$}
    $x_i = x_c' \odot I_{i-1}$\tcp*{keep results for those will be clipped in $\alpha_{i-1}$.}
    $x' += x_i$\tcp*{keep de-noised data into final output.}}

$x_1 = F_I\left(C\left(x, \alpha_1\right)\left|\right.\theta\right)$\tcp*{Clip \& De-noise the least range $\alpha_1$}
$x' += x_1 \odot \left([1]^{m \times n} - I_t- I_{t-1} - \cdots - I_1\right)$\tcp*{keep residual data into output}
\Return{$x'$}
 \caption{N2N-Seismic: Clip \& De-noise}
\end{algorithm}

\subsection{Evaluation Metrics}

To evaluate the performance, we use the following measurements. The signal-to-noise-ratio (SNR) is defined as the ratio between the variance of the original gather and of the noise, where noise is the difference between the corrupted signal and the clean signal. Given a corrupted seismic data $S$ (or denoised seismic $S'$), and its clean sample $\hat{S}$, SNR (signal-to-noise-ratio) is defined as:

\begin{equation}
    SNR = 10\times log_{10} \frac{(\hat{S})^2}{(S-\hat{S})^2}
    \label{equation:snr}
\end{equation}

Mean-Squared-Error (MSE) is defined as the average of the element-wise squared difference between the predicted signal and true signal (both of them with the size of $M \times N$), calculated as follows:

\begin{equation}
    MSE =  \frac{1}{M \times N}\sum_{i=1}^{M} \sum_{j=1}^{N}  \left| \left|S_{i,j} - \hat{S}_{i,j}\right|\right|^2.
    \label{equation:mse}
\end{equation}

The aforementioned measurements are wildly used in traditional image attenuation evaluation. However, in seismic data, we should also pay attention to the amplitude, phase, and correlation coefficients between predicted signal and original signal. We would like to show such evaluation measurements in the next section visually.


\section{Case Study 1: Wedge Model Data}
\label{sec:wedge}
\subsection{Wedge Dataset Preparing}
In this experiment, a simple wedge model is generated with dimension of 200 samples and 51 traces. Then, 4 different levels of random noise are added to the wedge model, where the Gaussian white noise with mean $\mu = 0 $ and scale $\sigma = 0.01$, $0.03$, $0.07$, and $0.10$, respectively. Data value has been normalized to $[-1,1]$ range. Figure \ref{fig:wedge_comparison_image} column (a) shows the noise-added seismic data, indicating the information of the wedge model. 

\subsection{Experimental Settings}
For model training, rather than using the seismic image, we used the image dataset on ImageNet\footnote{Dataset available at \url{http://image-net.org/download-imageurls}}. We randomly select 300 images for training and 50 images for validation. Those images consist of different categories, such as animals, plants, landscapes, etc.

As we described in the previous section, we used the idea from Noise2Noise model and added random Gaussian noise to the input and output images, separately, during the training processes. We applied the loss function in Eq. \ref{equation:noise2noise_loss}. At the meanwhile, MSE (Eq. \ref{equation:snr}) and SNR (Eq. \ref{equation:mse}) would be used as the evaluation metrics.

Training process would be automatically terminated once the performance converges. All parameter weights would be saved to file. For the hyper-parameters tuning, we applied the validation data to adjust hyper-parameters, i.e., learning rate, feature dimension, number of the residual unit, and steps per epoch. We adopt the Adaptive Moment Estimation  
\begin{wraptable}{r}{8cm}
\centering
\begin{spacing}{1.5}
\small
\begin{tabular}{c|c}
\toprule[2pt]
  Hyper-Parameter & Optimal Value\\ \midrule[1pt]
  \textit{Learning Rate} & 0.01\\ \hline
  \textit{Feature Dimension} & 64\\ \hline
  \textit{\# of Residual unit} & 16\\ \hline
  \textit{Optimizer} & Adam\\ \hline
  \textit{Step per epoch} & 1000  \\\bottomrule[2pt]
\end{tabular}
\vspace{-30pt}
\caption{Optimal Hyper-parameters Setting}
\label{tab:hyper-parameter}
\end{spacing}
\end{wraptable}
(Adam: \cite{kingma2014adam}) as the optimizer for training since it yields faster convergence compared to Stochastic Gradient Descent (SGD). The optimal hyper-parameters (listed in Table \ref{tab:hyper-parameter}) would be used into the final model training.

Next, we tested the trained DL model on the wedge model data. As we discussed in the previous section, the proposed DL model, named N2N-Seismic, would apply the clipping process and apply on different ranges of the data value (recall Eq. (\ref{equation:clipping_and_applying}). The number of iterations, $t$, depends on the distribution of the data value. Since all data value in the wedge dataset relatively uniform distributed from -1 to 1, we chose a low number of iterations, $t = 2$, for the final prediction. Therefore, the threshold $\alpha$ would be set as 0.5 and 1, respectively.

\subsection{Results and Analysis}
Landmark Solutions \footnote{Landmark Solutions: \url{https://www.landmark.solutions/}} is an E\&P software widely used for seismic data processing (Abbr. \textit{Reference}). It adopts the FX-Decon algorithm for random seismic noise attenuation. In this manuscript, we would like to use their solution as a rigorous benchmark for comparison.

\begin{figure*}[htbp]
    \centering
    \includegraphics[width =\textwidth]{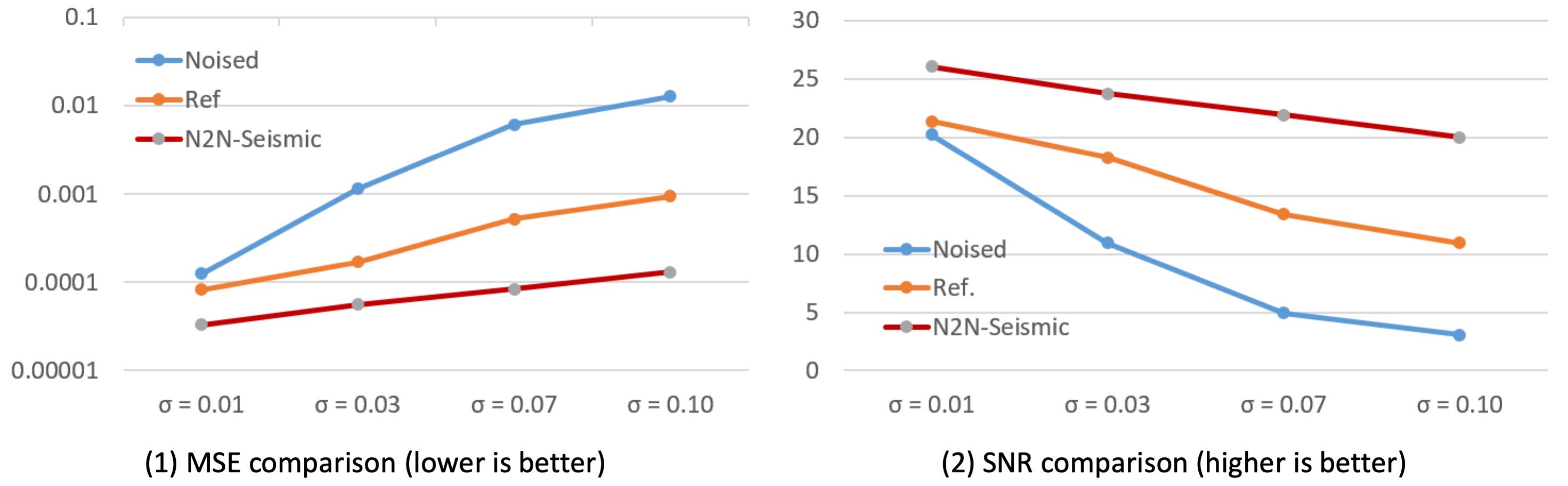}
    \vspace{-30pt}
    \caption{Performance comparisons on the Wedge dataset in different noise levels.}
    \label{fig:wedge_mse_snr}
\end{figure*}

\begin{figure*}
\centering
  \includegraphics[width=\textwidth]{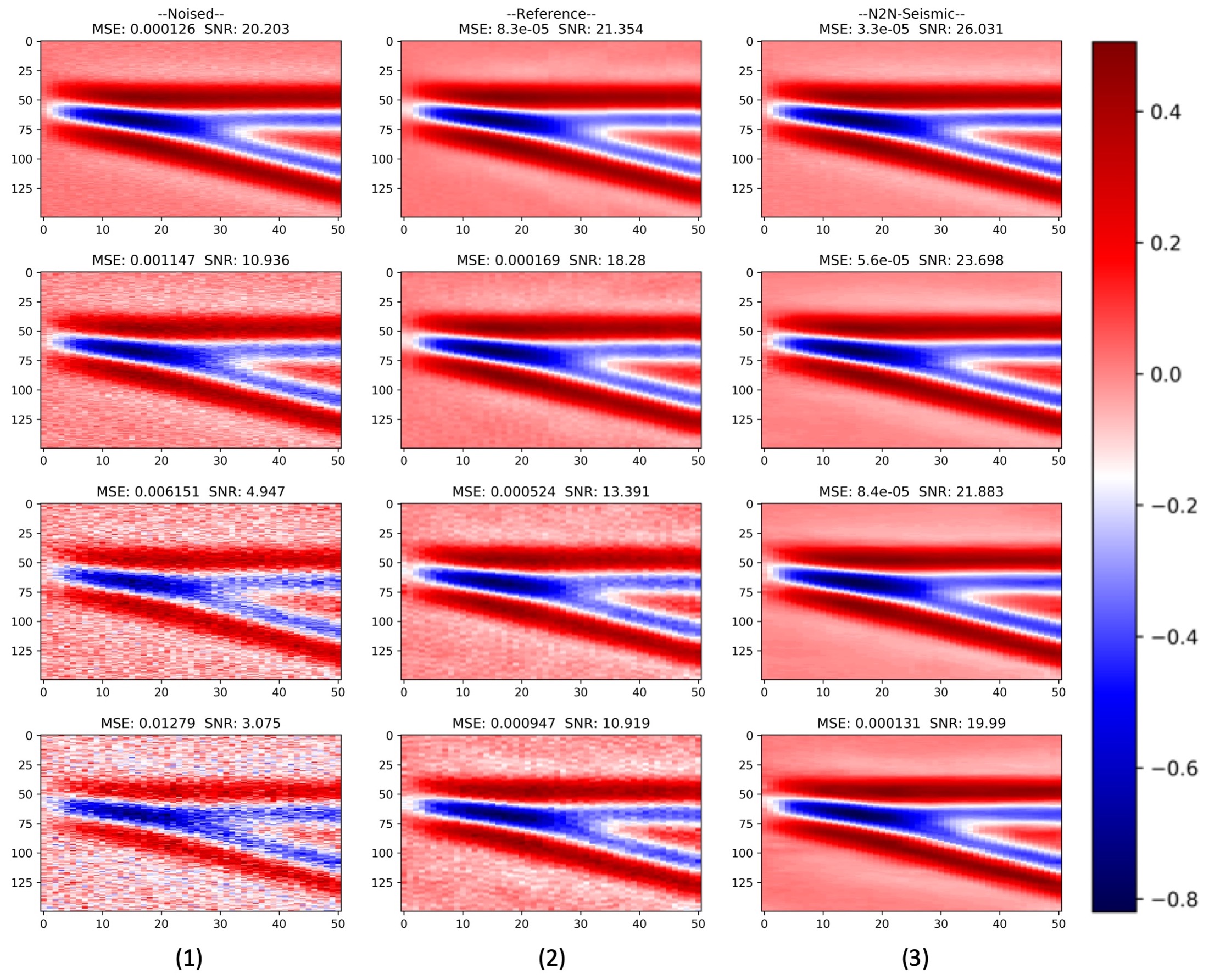}
  \vspace{-50pt}
  \caption{Random noise attenuation for wedge model data. Column (1) displays the noise-added seismic data with different level of Gaussian noise; Column (2) shows the de-noised results from reference methods; and Column (3) shows the de-noised results from N2N-Seismic Model.}
  \label{fig:wedge_comparison_image}
\end{figure*}

Figure \ref{fig:wedge_comparison_image} shows the random noise attenuation results for the wedge model data. Column (1) displays the noise-added seismic data with different levels of Gaussian noise. Comparing with the ground true clean data (not shown), as we can see, the MSE of noised data increases from $1.26 \times 10^{-4}$ to $1.28 \times 10^{-2}$, with the growth of noise level (from top to bottom). In addition, the SNR value has decreased from 20.2 dB to 3.1 dB with the growth of noise level. We would use these corrupted seismic data as the input of our N2N-Seismic Model for noise attenuation.

\subsubsection{Noise-level Sensitivities} Figure \ref{fig:wedge_comparison_image} Column (2) shows the de-noised results from reference methods. From the perspective of evaluation metrics, both MSE and SNR of these de-noised images are highly noise-level sensitive. Specifically, the ability of noise attenuation would be sharply decreased with the growth of noise level, from $(MSE = 8.30 \times 10^{-5}, SNR = 21.4\ dB)$ to $(MSE = 9.47 \times 10^{-4}, SNR = 10.9\ dB)$. Figure \ref{fig:wedge_comparison_image} Column (3) shows the de-noised results from our N2N-Seismic Model. Comparing with Reference model, the ability of noise attenuation of N2N-Seismic has a mildly decreased with the growth of noise level, from $(MSE = 3.30 \times 10^{-5}, SNR = 26.0\ dB)$ to $(MSE = 1.31 \times 10^{-4}, SNR = 20.0\ dB)$. This observation indicates that N2N-Seismic model is more robust and less noise-level sensitive than the conventional methods.

\subsubsection{De-noising Abilities} The evaluation metrics, MSE and SNR, show that our N2N-Seismic model has significantly better performance than the reference method. Specifically, in the Level-1 noise case, MSE is dropped from $1.26 \times 10^{-4}$ to $3.30 \times 10^{-5}$. Comparing with reference model ($MSE = 8.3 \times 10 ^{-5}$), our method has 62.50\% improvement. And SNR has been improved to 26.0 dB, comparing with the reference model ($SNR = 21.4\ dB$), N2N-Seismic model has improved 21.50\%. Moreover, such improvement would be more apparent with the growth of noise level. For example, in the Level-4 case, N2N-Seismic has improved 86.3\% on MSE and 75.3\% on SNR, compared to the reference method. Such results indicate that deep-learning-based N2N-Seismic model performs much better than the conventional methods and makes an impressive achievement.

\subsubsection{Signal Recovery Abilities} 
\begin{figure}
\centering
  \includegraphics[width=0.8\textwidth]{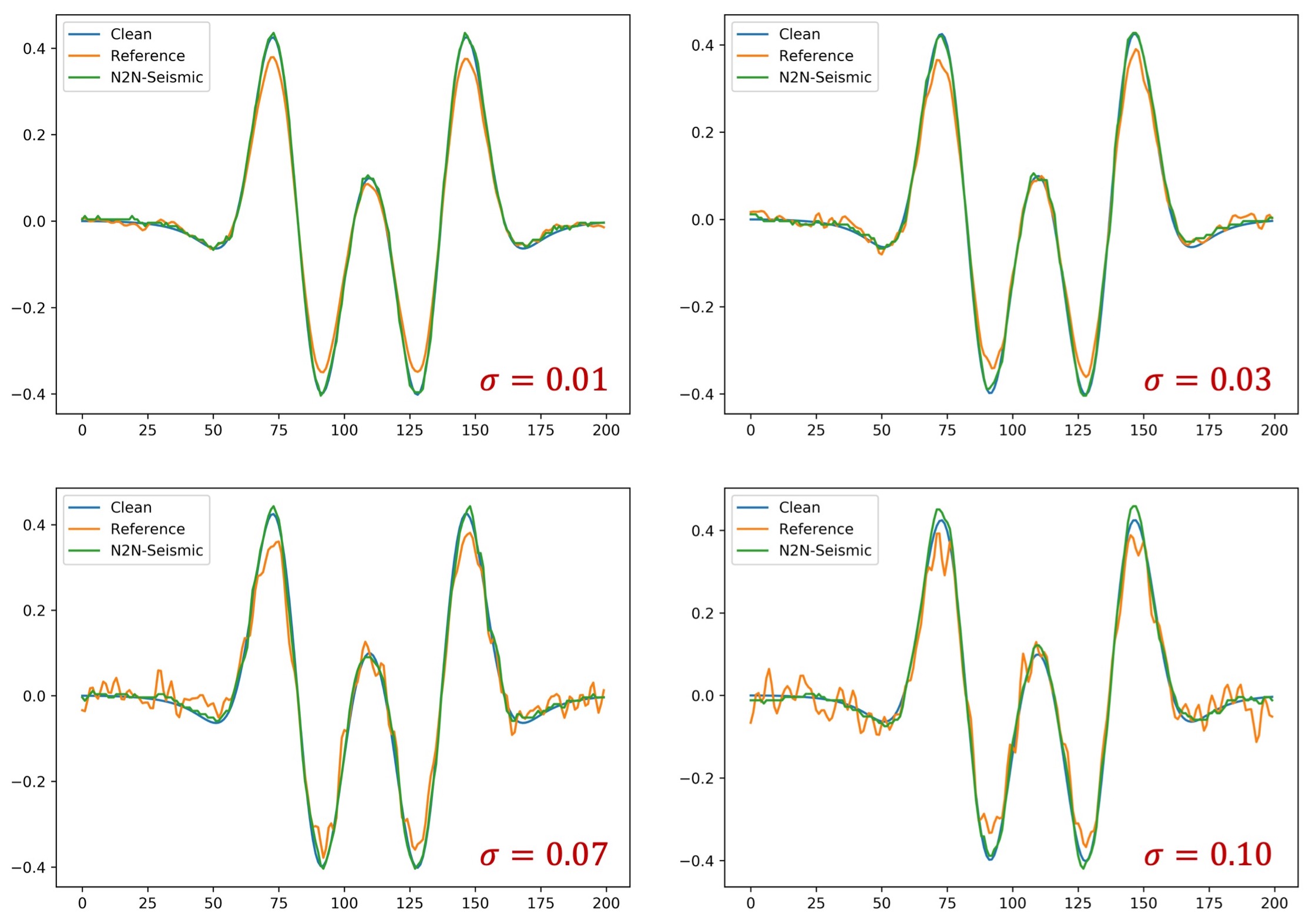}
  \vspace{-30pt}
  \caption{An example of signal recovery using N2N-Seismic and reference method in different noise levels.}
  \label{fig:wedge_comparison_signal}
\end{figure}

Unlike ordinary image processing, seismic data processing requires more strict criteria in respect of phrase, amplitude, etc., during signal recovery. Figure \ref{fig:wedge_comparison_signal} shows a randomly selected trace from the wedge model, recovered trace by the reference method and N2N-Seismic model. N2N-Seismic recovered signal (blue) is much closer to the ground truth (green) than the recovered signal by the reference method (yellow), in all noise levels. Specifically, N2N-Seismic recovered signals have similar phase and amplitude as the clean signal; however, reference method recovered signal would always have the lower range of amplitude, and its high-frequency residual noise would change the phase of the original clean signal, especially in the high noise-level cases. This observation once again indicates N2N-Seismic model has stronger abilities of seismic noise attenuation and signal recovery than the conventional method.

\subsection{Summary}

In this case study, we apply our N2N-Seismic model onto synthetic Wedge dataset and obtain promising results. N2N-Seismic effectively 1) reduces the MSE with evaluation metric of 74.5\% in average of 4 noise levels cases; 2) improves the SNR evaluation metric with about 44.0\% on average of 4 noise level cases; 3) keeps the original phase and amplitude information of the original clean signal; and 4) reduces the effects of noise level with recovered signals. This study case on wedge data is comparably less challenging for robustness validation of the N2N-Seismic model due to the relatively uniform data distribution. In the next section, we will test the performance with a more complicated case.

\section{Case Study 2: SEAM Data}
\label{sec:seam}
In the previous section, we applied N2N-Seismic model onto the wedge data and got the promising results. However, this task is relatively simple since the wedge model has uniform data distribution. In this section, we apply the N2N-Seismic model into a more complicated situation, where the data distribution is much more intricate than the wedge model data. Therefore, we will see all the benefits of N2N-Seismic model, especially from the Eq. \ref{equation:clipping_and_applying}. The performances will not only be compared with the conventional method, but also compared with the deep learning methods, N2N-Image (Abbr. N2N-I) directly onto the seismic data.

In this case study, we use the public SEG Advanced Modeling Program (SEAM) data\footnote{SEAM data available at \url{https://seg.org/News-Resources/Research-and-Data/SEAM}}. A single line on this dataset is randomly selected. The dataset has 1600 samples in depth with a sample rate of 5 meters, which covers a total depth of 8 kilometers from water surface; and the selected testing area contains 245 traces. Figure \ref{fig:seam_whole_comparison} column (1) shows the clean seismic data (horizontally compressed for saving space). 

\subsection{SEAM Dataset Preparing}
\begin{wrapfigure}{r}{9cm}
\includegraphics[width=9cm]{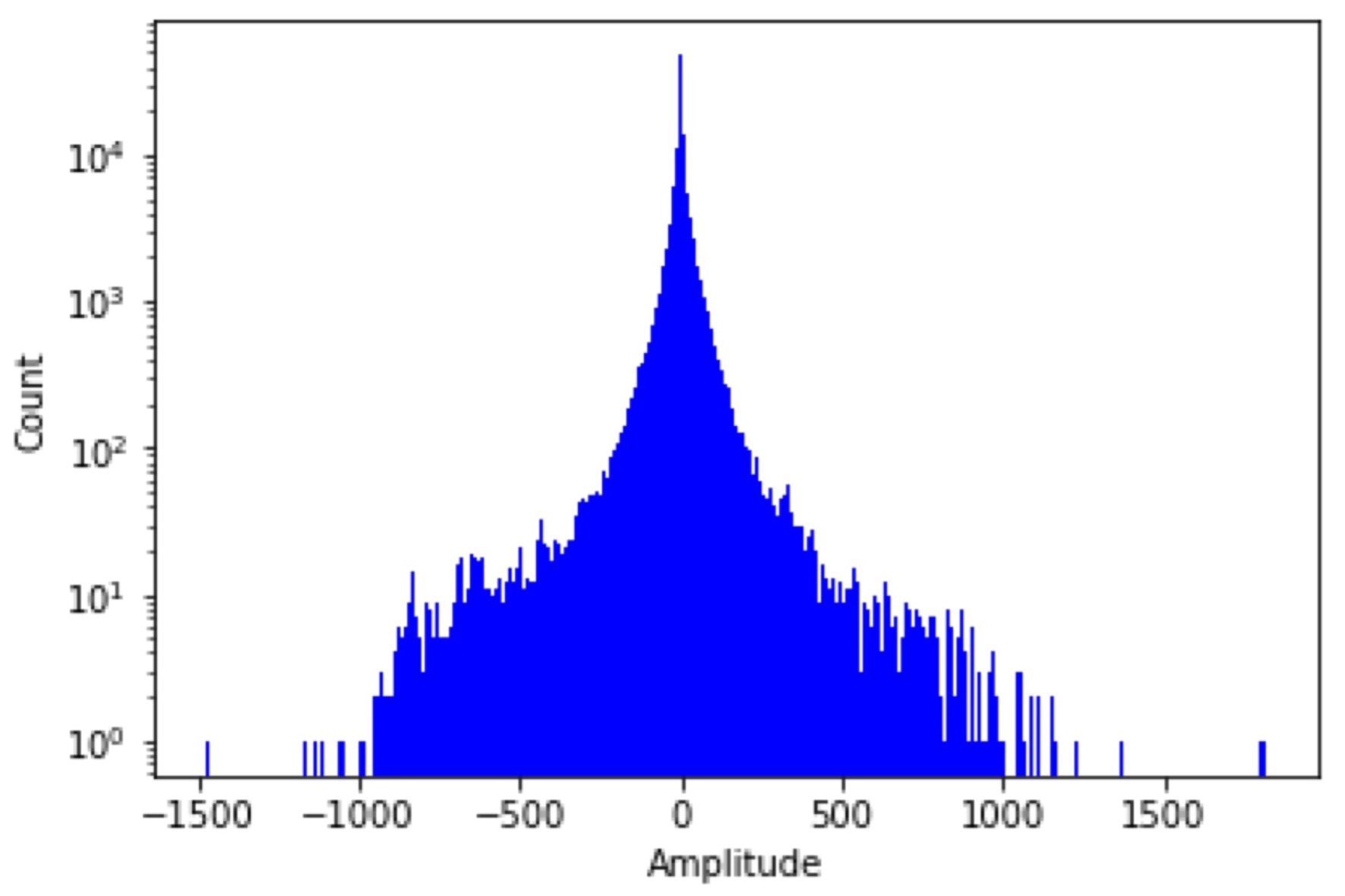}
\vspace{-50pt}
\caption{Amplitude Distribution on SEAM data}
\label{fig:seam_data_distribution}
\end{wrapfigure}

Like wedge model case study, we have added different (3) levels of random noise under the water bottom line. Denote $A$ as the max absolute amplitude and $\mu_0$ is the mean of the SEAM data. The added Gaussian white noise with mean $\mu = \mu_0 $ and scale $\sigma = 0.03 \times A$, $0.09\times A$, and $0.15\times A$, respectively. We did not perform the normalization onto the SEAM data. Figure \ref{fig:seam_whole_comparison} column (2), (5), and (8) show the noise-added seismic data.

However, unlike the wedge dataset, the SEAM dataset has more complicated distribution. Figure \ref{fig:seam_data_distribution} shows the histogram of the data distribution. As we can see, most of the amplitudes from the data concentrate on a small amplitude range $[-200,200]$; therefore, for such majority, the generated noise would cover a large range of signal information due to the strong amplitude of noise. In conclusion, this recovery task is more challenging for the noise attenuation model.

\subsection{Experimental Settings}
Like the wedge model, we use the N2N-Seismic model trained with the ImageNet dataset. We also apply the loss function in Eq.  \ref{equation:noise2noise_loss}, and use MSE (Eq. \ref{equation:snr}) and SNR (Eq. \ref{equation:mse}) as evaluation metrics. All optimal hyper-parameters are listed on Table \ref{tab:hyper-parameter}. 

Next, we apply our SEAM Phase I data (\cite{fehler2011seam}) using the trained model with optimal parameters. As we discussed on the previous section, N2N-Seismic, would need to perform clippings with different range of data value (recall Eq. \ref{equation:clipping_and_applying}). The number of iterations, $t$, depends on the data value distribution. Since the data value in the SEAM dataset has broader range, we chose higher number of iterations, $t = 5$, for the final prediction, and the threshold $\alpha$ would be evenly set as $\alpha_1 = 0.2 \times A$, $\alpha_2 = 0.4 \times A$, $\alpha_3 = 0.6 \times A$, $\alpha_4 = 0.8 \times A$, and $\alpha_5 = 1.0 \times A$, respectively.

For the pre-trained model where input data has been normalized, the SEAM data is also normalized by dividing each threshold $\alpha_t$. Then, the pre-trained model is applied for noise attenuation, and the result is marked as $F_I(C(x, \alpha_t))$. Then we do the element-wisely producing the predicted value and the binary mask matrix $I_t$, and recovery the normalized data by multiplying the threshold $\alpha_t$. Finally, the noise attenuated data is generated by integrating with $t$ iterations of inferences.

\subsection{Results and Analysis}

\begin{figure*}
\centering
  \includegraphics[width=\textwidth]{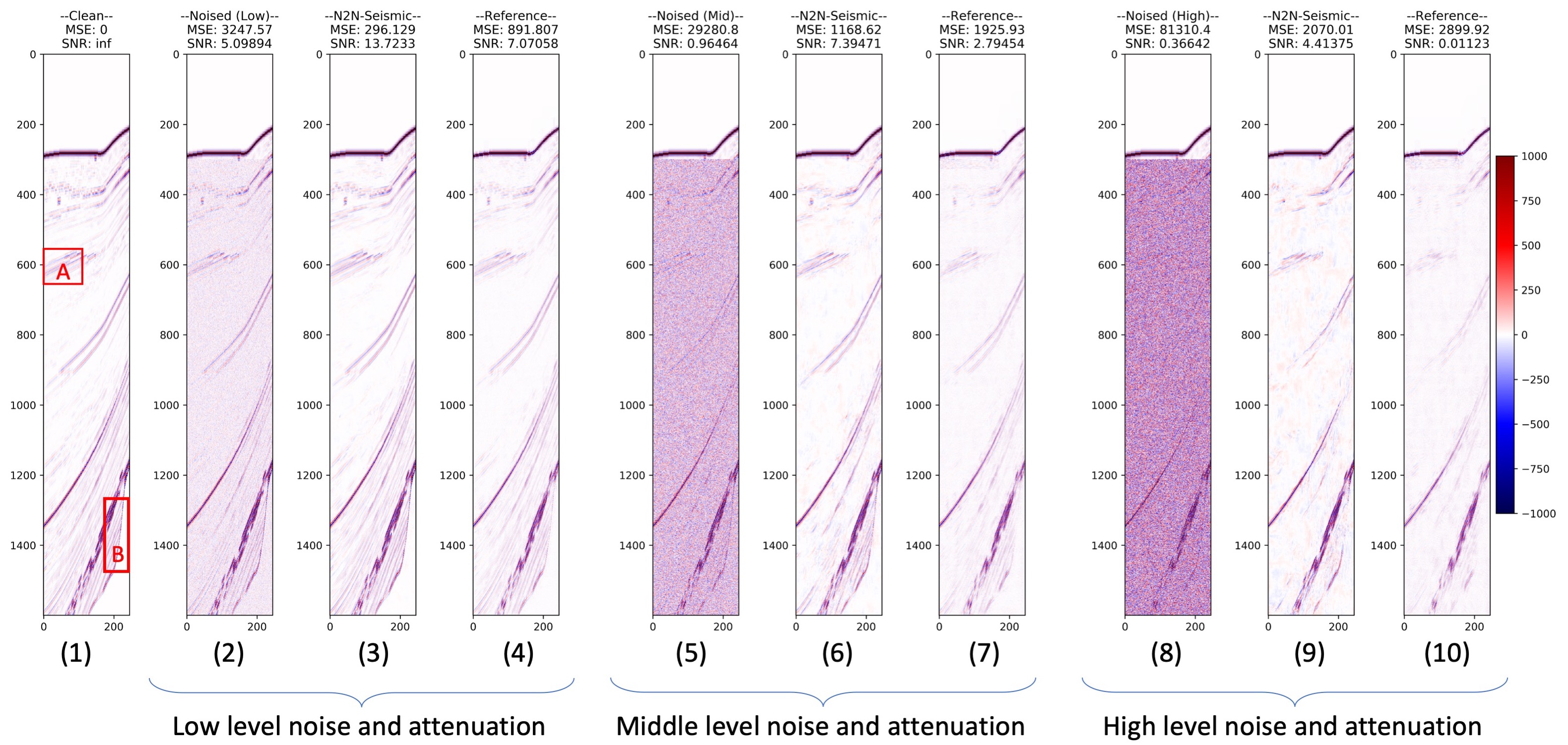}
  \vspace{-50pt}
  \caption{Performance comparisons for random noise attenuation. Figure (1) shows a clean image without random noise. Figure (2), (5), and (8) show the clean image with low, middle, and high-level random noise. Figure (3), (6), and (9) are the de-noised seismic data by our N2N-Seismic method, and Figure (4), (7), and (10) are the de-noised seismic data by the reference model.}
  \label{fig:seam_whole_comparison}
\end{figure*}

\begin{figure*}
\centering
  \includegraphics[width=\textwidth]{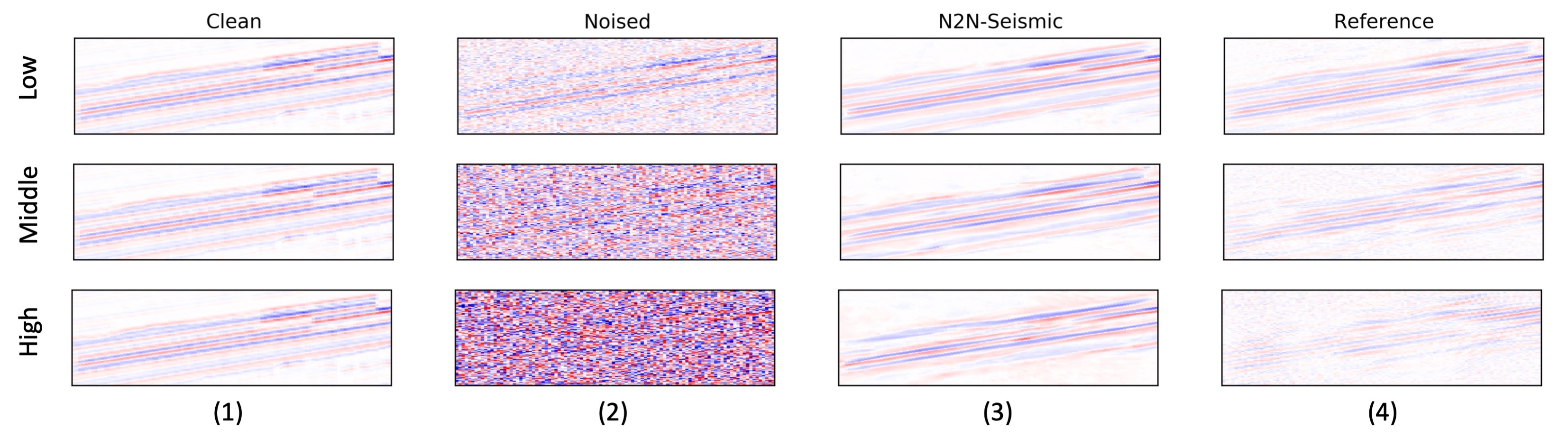}
  \vspace{-30pt}
  \caption{Performance comparison for random noise attenuation (Zoomed Part A). Figure column (1) shows the partial clean image without random noise. Figure column (2) shows the partial clean image with low, middle, and high-level random noise. Figure column (3) is the de-noised seismic data by our N2N-Seismic methods, and Figure column (4) is the de-noised seismic data by the reference model.}
  \label{fig:seam_partial_comparison_A}
\end{figure*}

\begin{figure*}
\centering
  \includegraphics[width=0.85\textwidth]{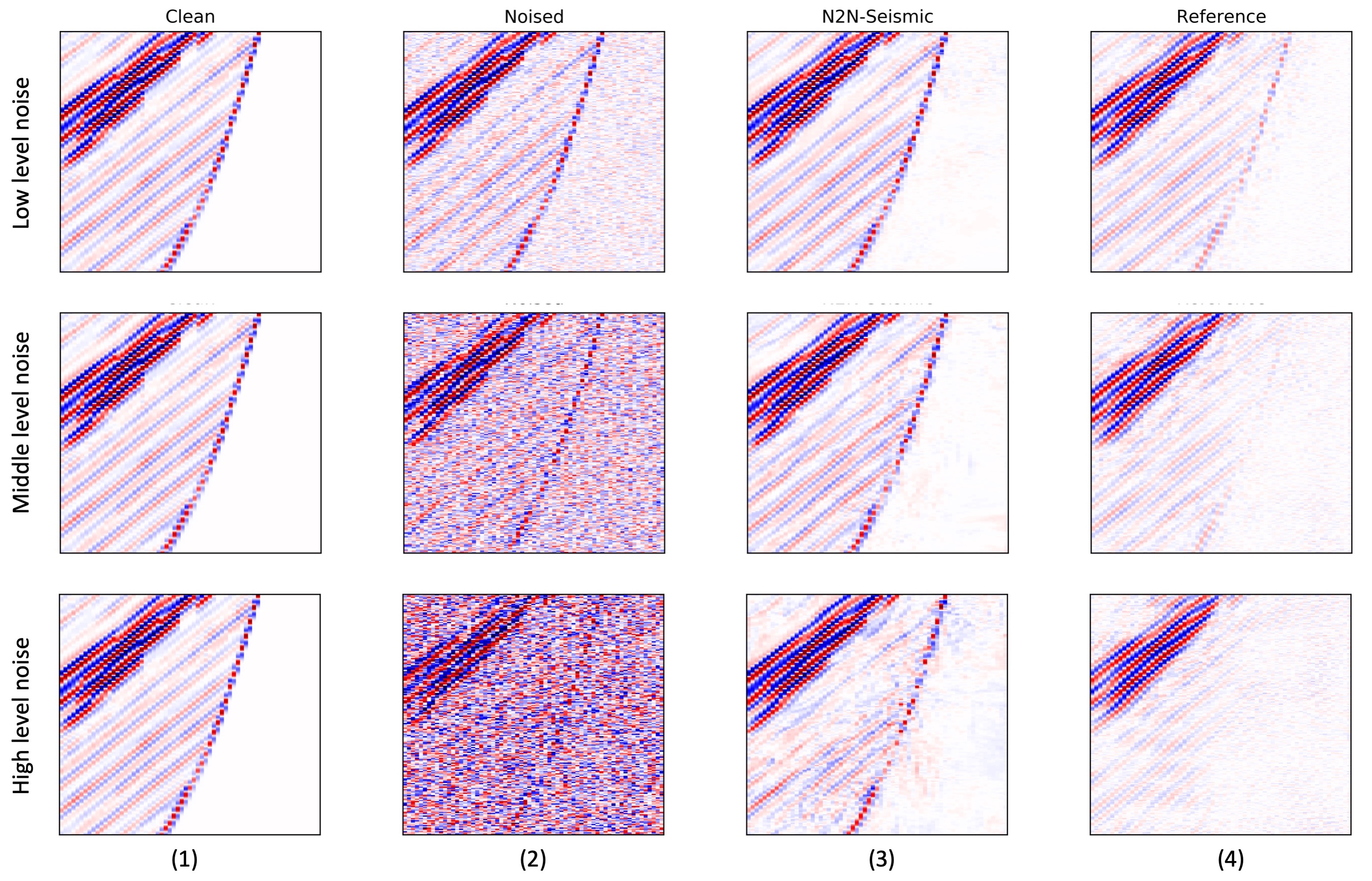}
  \vspace{-30pt}
  \caption{Performance comparisons for random noise attenuation (Zoomed Part B). Figure column (1) shows the partial clean image without random noise. Figure column (2) shows the partial clean image with low, middle, and high-level random noise. Figure column (3) is the de-noised seismic data by our N2N-Seismic methods, and Figure column (4) is the de-noised seismic data by the reference model.}
  \label{fig:seam_partial_comparison_B}
\end{figure*}

Figure \ref{fig:seam_whole_comparison} exhibits the comparison of performances for random noise attenuation using the conventional method and our N2N-Seismic model. Figure \ref{fig:seam_whole_comparison} (1) shows the clean image without random noise. Figure \ref{fig:seam_whole_comparison}  (2), (5), and (8) sshow the clean image with low, middle, and high-level random noise under the water bottom line. With the growth of the noise level, we observe that the primary signals are gradually covered by the noise. Refer to Figures \ref{fig:seam_partial_comparison_A} and \ref{fig:seam_partial_comparison_B}, which are partial data zoomed from the red quadrangle regions A and B on Figure \ref{fig:seam_whole_comparison} (1), we could observe that the primary signals and many details of seismic data have been occupied by the Gaussian noise (comparing (1) and (2)). The primary signals have lower absolute amplitude than noise, as we discussed before. Therefore, recovering such signal information from noises becomes more challenging than the previous case study.

As a comparison, Figure \ref{fig:seam_whole_comparison} (3), (6), and (9) show the de-noised seismic image by N2N-Seismic model. Comparing with the results from the reference method, N2N-Seismic model recovers more details with much cleaner results, for both low and high absolute amplitude regions. Zooming to the selected quadrangle region, referring to Figure \ref{fig:seam_partial_comparison_B} (3), more details have been recovered, especially for the bottom edge.

Figure \ref{fig:seam_whole_comparison}  (4), (7), and (10) exhibit the de-noised results by the reference methods. As we can see, the signals with high absolute amplitude have been recognized and covered; however, the recovered seismic data loses much more details compared with the ground truth. More specifically, referring to the close-up image in Figure \ref{fig:seam_partial_comparison_A} and Figures \ref{fig:seam_partial_comparison_B}, with the growth of the noise level, de-noised images by the reference method lose more details, especially at the bottom edge.

We have initially determined that N2N-Seismic model has a stronger ability for noise attenuation than the conventional method. In the following sub-sections, we will evaluate the results with more detailed numeric comparison, respect to MSE, SNR, correlation coefficient, phase, etc. In the following comparison, we will also include the traditional deep learning method which directly applied to seismic data (named N2N-Image), for manifesting the performance of our proposed model, N2N-Seismic.

\subsubsection{MSE and SNR}

\begin{table}[]
\centering
\begin{tabular}{c|c|c|c|c|c}
\toprule[2pt]
                      &          & Noised   & Ref      & N2N-I    & N2N-S    \\ \midrule[1pt]
\multirow{3}{*}{low}  & MSE      & 3247.576 & 891.807  & 460.695  & \textbf{296.129}  \\ \cline{2-6} 
                      & SNR      & 5.098    & 7.070    & 11.953   & \textbf{13.723}   \\ \cline{2-6} 
                      & CorrCoef & 0.7936   & 0.9430   & 0.9622   & \textbf{0.9743}   \\ \midrule[1pt]
\multirow{3}{*}{mid}  & MSE      & 29280.86 & 1925.938 & 1323.412 & \textbf{1168.622} \\ \cline{2-6} 
                      & SNR      & 0.964    & 2.794    & 7.011    & \textbf{7.394}    \\ \cline{2-6} 
                      & CorrCoef & 0.4225   & 0.8463   & 0.8855   & \textbf{0.8954}   \\ \midrule[1pt]
\multirow{3}{*}{high} & MSE      & 81310.40 & 2899.924 & 2152.178 & \textbf{2070.015} \\ \cline{2-6} 
                      & SNR      & 0.366    & 0.011    & 4.366    & \textbf{4.413}    \\ \cline{2-6} 
                      & CorrCoef & 0.2703   & 0.7460   & 0.8076   & \textbf{0.8126}   \\ \bottomrule[2pt]
\end{tabular}
\vspace{-30pt}
\caption{Performance comparisons in terms of MSE, SNR, and Correlation Coefficient. The best results are labeled as bold.}
\label{tab:SEAM_performance}
\end{table}

Table \ref{tab:SEAM_performance} shows the comparison regarding the performance between the reference method and N2N-Seismic. In the low-level noise case, the MSE of the de-noised results by N2N-Seismic model decreased from 3247.6 (noised image) to 296.1 (90.9\% attenuated); Comparing with the reference results ($MSE = 891.8$), we have improved 66.8\%; and comparing with the N2N-Image results  ($MSE = 460.7$), we have improved 35.7\%. In terms of the SNR, N2N-Seismic result ($SNR = 13.7\ dB$) improved 168.6\% from the noised data, significantly improved 93.0\% from the reference result ($SNR = 7.1\ dB$), and also improved 14.2\% from the N2N-Image result ($SNR = 12.0\ dB$).

Similar improvement could be found in the middle noise-level case. The MSE of the de-noised results by N2N-Seismic model decreased from 29280.9 (noised image) to 1168.6 (96.0\% attenuated); comparing with the reference results ($MSE = 1925.9$), we have improved 39.32\%; and comparing with the N2N-Image results ($MSE = 1323.4$), we have improved 11.7\%. In terms of the SNR, N2N-Seismic result ($SNR = 7.4\ dB$) improved 667.0\% from the noised data ($SNR = 1.0\ dB$), significantly improved 164.6\% from the reference result ($SNR = 2.8\ dB$), and improved 5.6\% from the N2N-Image result ($SNR = 7.0\ dB$).

More pronounced improvement could be observed in the high noise-level case. The MSE of the de-noised results by N2N-Seismic model decreases from 81310.4 (noised image) to 2070.0 (97.5\% attenuated); comparing with the reference results ($MSE = 2899.9$), we have improved 28.6\%; and comparing with the N2N-Image results ($MSE = 2152.2$), we have improved 3.8\%. In terms of the SNR, N2N-Seismic result ($SNR = 4.41$) improved 1105.7\% from the noised data, and dramatically improved more than 400 times from the reference result ($SNR = 0.011\ dB$) as well as 1.1\% from the N2N-Image result ($SNR = 4.36\ dB$).

It should be highlighted here that the reference method has an extremely low performance on improving the SNR results. In some cases, e.g., high-level noise, the de-noised results by reference model even have lower SNR than the noised image, which indicates conventional methods have limited ability on seismic noise attenuation in high-level noise case. As an alternative solution, N2N-Seismic gets much better results with respects to MSE and SNR; Comparing with N2N-Image model, which directly applies traditional deep learning model on seismic data, conventional methods still could not provide satisfactory solution.

\subsubsection{Correlation Coefficient}

\begin{table}[]
\centering

\begin{tabular}{c|c|c|c|c|c|c|c}\toprule[2pt]
      Noise           & Freq. (Hz)  & 0-10   & 10-20    & 20-30     & 30-40    & 40-50  & 50-60\\\midrule[1pt]
\multirow{3}{*}{low}  & Ref         & 0.938  & 0.956    & 0.984     & \textbf{0.991}    & 0.047  & \textbf{0.147}  \\
                      & N2N-I       & 0.872  & 0.789    & 0.989     & 0.985    & 0.091  & -0.076 \\
                      & N2N-S       & \textbf{0.946}  & \textbf{0.973}    & \textbf{0.993}     & 0.987    & \textbf{0.397}  & -0.461  \\\midrule[1pt]

\multirow{3}{*}{mid}  & Ref         & 0.929  & 0.916    & 0.983     & 0.976    & -0.21  & -0.61    \\
                      & N2N-I       & 0.907  & 0.939    & 0.980     & 0.969    & 0.242  & 0.218  \\
                      & N2N-S       & \textbf{0.947}  & \textbf{0.964}    & \textbf{0.985}     & \textbf{0.995}    & \textbf{0.040}  & \textbf{0.496}    \\\midrule[1pt]

\multirow{3}{*}{high} & Ref         & 0.845  & 0.915    & 0.978     & 0.986    & 0.056  & 0.233   \\
                      & N2N-I       & 0.890  & 0.910    & 0.979     & 0.988    & 0.119  & 0.337  \\
                      & N2N-S       & \textbf{0.939}  & \textbf{0.938}    & \textbf{0.982}     &  \textbf{0.991}    & \textbf{0.279}  & \textbf{0.404}  \\\bottomrule[2pt]
\end{tabular}

\vspace{-30pt}
\caption{Correlation Coefficients of Phase Spectrum. The best results are labeled as bold.}
\label{tab:corrcoef_phase}
\end{table}

Unlike ordinary image processing, in the geophysical field, the correlation coefficient between the de-noised signals and the original clean signals are extremely important because it reveals the consistency of seismic phase. Therefore, we consider the correlation coefficient as another evaluation metric in Table \ref{tab:SEAM_performance}. Due to the added Gaussian noise, the correlation coefficient between noised data and the ground truth have been dropped to 0.79, 0.42, and 0.27 in different noise-levels, respectively. For the de-noised results by the N2N-Seismic model, the correlation coefficients have increased to 0.97, 0.90, and 0.81. Specifically, comparing with the conventional method,  in the low noise level case, we improved  3.3\% from the reference method ($CorrCoef = 0.943$) and 1.3\% from the N2N-Image model ($CorrCoef = 0.962$); in the middle noise level case, we improved 5.8\% from the reference method  ($CorrCoef = 0.846$) and 1.1\% from the N2N-Image model ($CorrCoef = 0.886$); and in the high noise level case, we improved 8.9\% from the reference method ($CorrCoef = 0.746$) and 0.6\% from the N2N-Image model ($CorrCoef = 0.808$).

\subsubsection{Phase Recovery}

\begin{figure*}
\centering
  \includegraphics[width=0.95\textwidth]{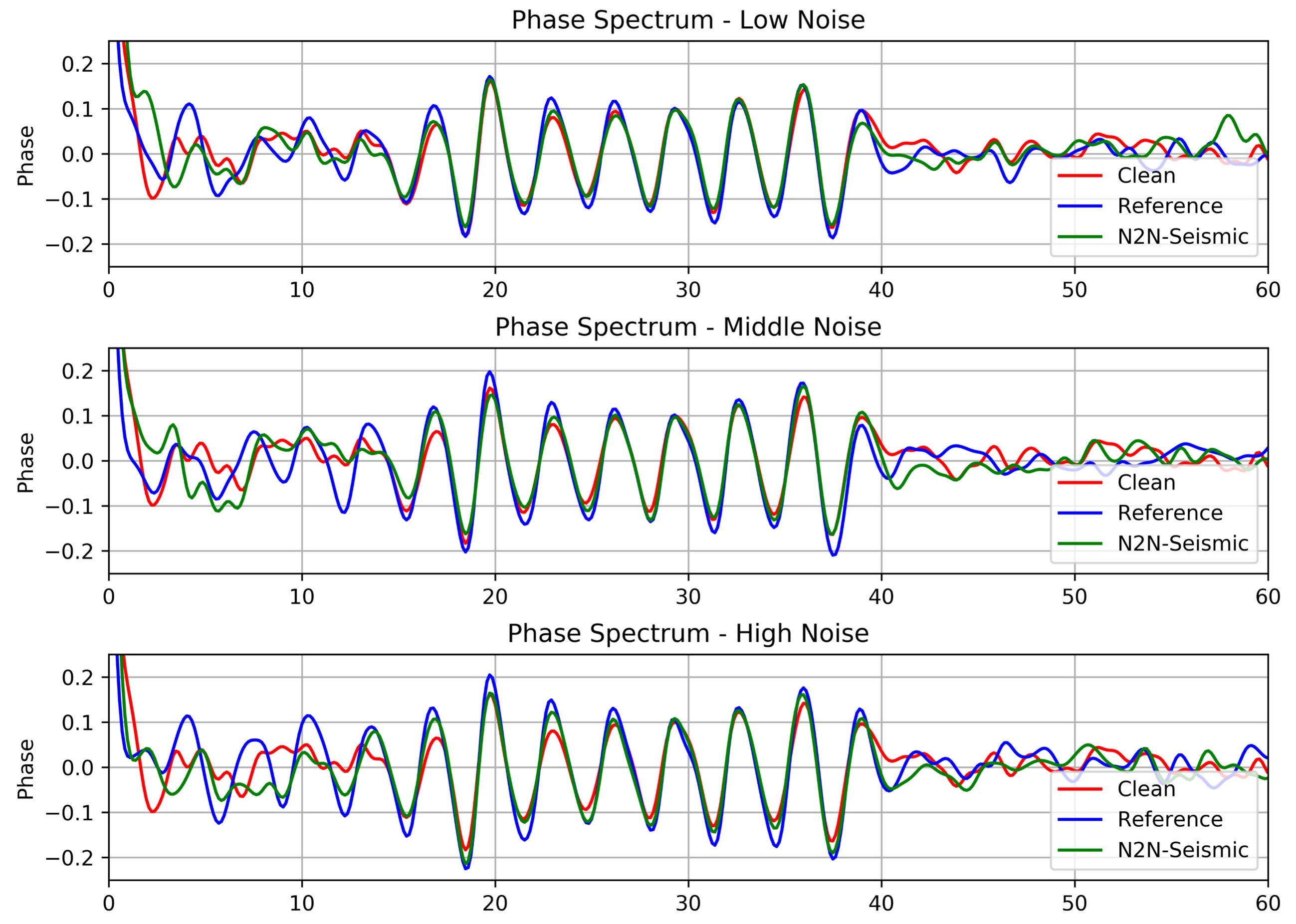}
  \vspace{-50pt}
  \caption{Phase spectrum comparison for random noise attenuation.}
  \label{fig:phase_comparison}
\end{figure*}


Next, we convert the dataset from depth to the time domain, and analyze the phase spectrum of de-noised data. Figure \ref{fig:phase_comparison} shows the phase spectrum of the original clean data and the de-noised data by the reference method and N2N-Seismic. We only show the phase spectrum between 0 Hz to 60 Hz. Because when frequency than 60 Hz, the amplitude of signal would be extremely low, so that such high-frequency components will not affect the primary signals and have fewer effects of follow-up processing and analysis. As we can see on Figure \ref{fig:phase_comparison}, frequency from 0 Hz to 19 Hz, results from N2N-Seismic are closer to the original phase than the ones from reference method in every level of noise; frequency after 19 Hz, results from N2N-Seismic would have perfectly fit the original phase as similar as results from N2N-Image and Reference methods. After 40 Hz, N2N-Seismic still perform better than the Reference model.


And Table \ref{tab:corrcoef_phase} shows the correlation coefficient of phase in important frequency range of 0 Hz to 60 Hz in details. Refer to Table \ref{tab:corrcoef_phase}, as we can see, for all frequency ranges, results from N2N-Seismic model are much better than results from the reference model in in all three noise level cases, except an outlier in Frequency 50 Hz to 60 Hz in low level case. Also, the correlation coefficient of phase from N2N-Seismic model is better than ones from N2N-Image model, which indicates that our model is more effective targeting to the seismic data. 



\subsubsection{Signal Information Recovery}

Finally, we would like to compare the abilities of signal recovery between the reference method, N2N-Image, and N2N-Seismic. As we discussed before, in the ordinary image processing using the deep learning model (N2N-Image), CNN has an objective of reducing the loss value, 
\begin{wrapfigure}{r}{11cm}
  \includegraphics[width=0.7\textwidth]{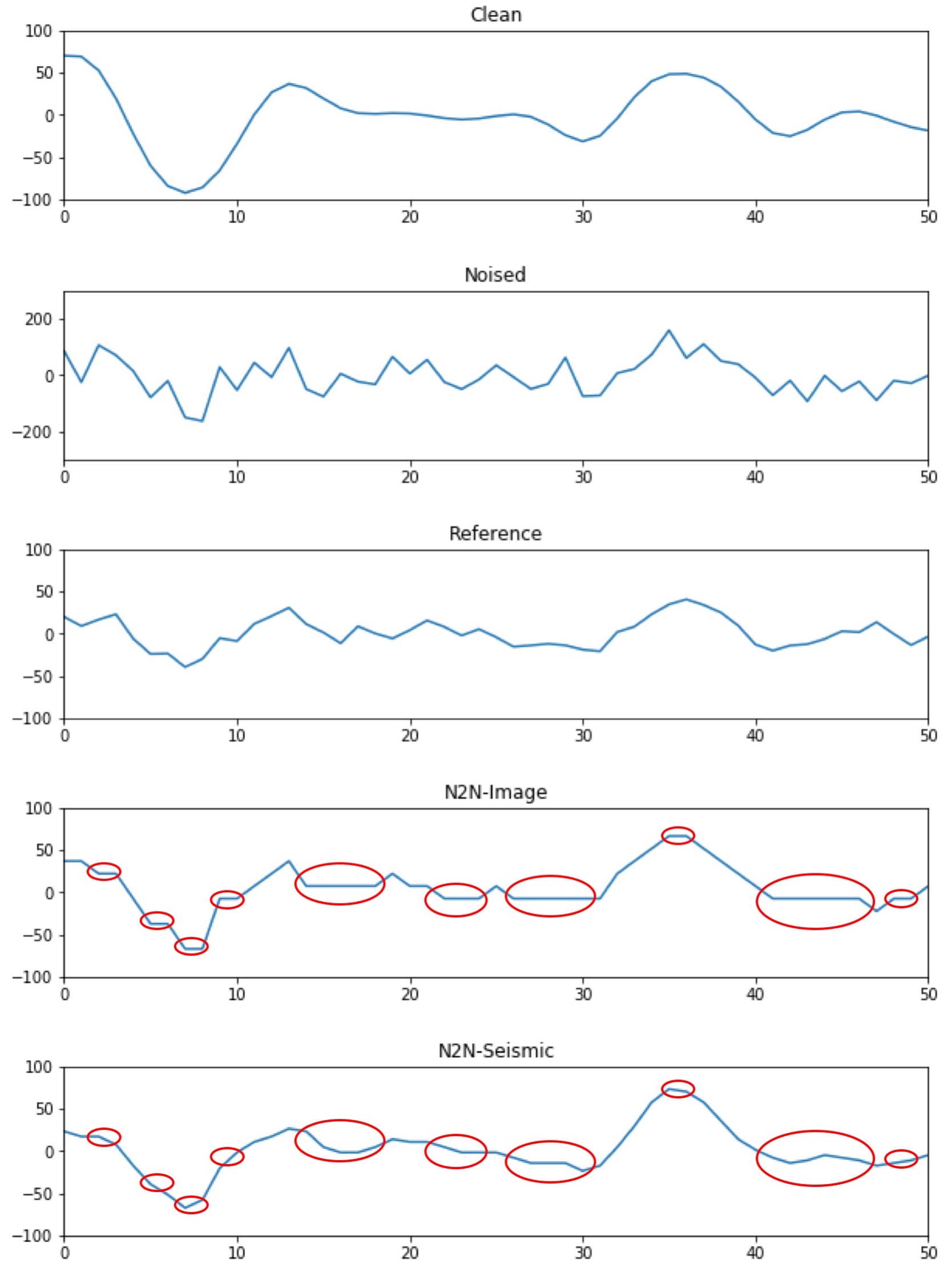}
  \vspace{-30pt}
  \caption{Wave comparison for random noise attenuation (high noise-level case).}
  \label{fig:signal_comparison}
\end{wrapfigure}
e.g., L1 loss or L2 loss, and make the predicted values converge to a certain level for training purpose. As a result, the de-noised images may exhibit less sharpen and looks smoother in the areas of adjacent pixels. Such results could not be acceptable in the geophysical field since the first priority of seismic attenuation is to keep the phase and amplitude spectrum of the signal close to intact. 

As an example, Figure \ref{fig:signal_comparison} shows a random trace in the depth range of 50 in the high noise level case. Comparing with the clean trace, the de-noised trace by N2N-image changes seismic phase circled as red. Such information loss would bring the big issue on seismic interpretation. As a comparison, the de-noised trace from the reference method recovers more phase information than N2N-Image, although it still remains some residual high-frequency noise. Reference methods for seismic noise attenuation would be a better solution than the traditional ordinary image processing method. However, the de-noised trace from N2N-Seismic convinces that it could be considered as a better approach to solve the information loss problem than the N2N-Image (refer to the red circles in same depth), and also apparently shows lower noise than the reference method.

\subsection{Summary}
In this case study, we exhibit the performance of N2N-Seismic model on a more complicated case. One the one hand, comparing with the conventional method, results from N2N- Seismic impressively improves the MSE and correlation coefficient value; furthermore, N2N-Seismic model dramatically increases the SNR of maximum 400 times. N2N-Seismic keeps the phase and amplitude spectrum as the original clean data, which proves to be much better than the conventional methods. However, for the phase correlation coefficient comparison, the conventional method did slightly better in the low frequency range.

On the other hand, comparing with the N2N-Image model, which directly applies the deep learning model designed for ordinary images onto the seismic data, N2N-Seismic still offers clear advantages in the respects of MSE, SNR, and correlation coefficient. Most importantly, N2N-Seismic keeps more details about the seismic information than N2N-Image, which would be much helpful for the subsequent data processing and analysis. These observations indicate our model, specifically designed for the seismic data, effectively achieves the objects of seismic noise attenuation.

\section{Conclusion}

In this manuscript, we proposed a deep learning model with CNN based residual neural networks for the random seismic noise attenuation tasks. Rather than directly applying de-noising model for the ordinary image to the seismic data, our proposed method, N2N-Seismic, has a strong ability in respect of recovery of the seismic wavelets back to intact condition while the signal is relatively preserved. Comparisons, from the two examples with wedge and SEAM data, show that our method performs much better than conventional methods for noise attenuation tasks in terms of SNR, MSE, and Phase Spectrum, etc.

In conclusion, the main contribution of this manuscript is to provide a deep-learning solution for random noise attenuation tasks. Such method absorbs benefits from the deep neural networks in computer vision applied to ordinary image denoising process, and meets the geophysical requirements and expectations. Having rigorous comparisons with conventional methods for several benchmark studies, our proposed deep-learning models successfully implement the tasks above with a great success, and achieve the prominent improvements in respects of MSE, SNR, etc..

\bibliographystyle{seg}  
\bibliography{ref}

\end{document}